\documentclass[letterpaper, 10 pt, conference, twocolumn, nobalancelastpage]{ieeeconf}  

\IEEEoverridecommandlockouts                              

\usepackage{graphics} 
\usepackage{graphicx}
\usepackage{subcaption}
\usepackage{float}
\usepackage{multicol}
\usepackage{booktabs}
\usepackage{tabularx}
\usepackage{algpseudocode} 
\usepackage{amsmath} 
\usepackage{amssymb}  
\usepackage{upgreek}
\usepackage{mathtools} 
\usepackage{esvect} 

\usepackage{kpfonts}

\usepackage{tikz}
\usetikzlibrary{shapes,arrows,shadows}
\usepackage{bm,times}
\makeatletter

\def\infinity{\rotatebox{90}{8}}

\usepackage{lettrine}

\title{\LARGE \bf
Robust Data Association for Object-level Semantic SLAM  
}

\author{Xueyang Kang$^{1}$ Shunying Yuan$^{2}$
\thanks{*This work was supported by Qualcomm R\&D Center, Beijing,China}
\thanks{$^{1}$Xueyang Kang is with robotic visual group at Qualcomm.
        {\tt\small xueykang@qti.qualcomm.com}}%
\thanks{$^{2}$Shunying Yuan is with  robotic visual group at Qualcomm.
        {\tt\small shunying@qti.qualcomm.com}}%
}

\begin{document}

\maketitle
\thispagestyle{empty}
\pagestyle{empty}

\begin{abstract}

Simultaneous mapping and localization (SLAM) in an real indoor environment is still a challenging task. Traditional SLAM approaches rely heavily on low-level geometric constraints like corners or lines, which may lead to tracking failure in textureless surroundings or cluttered world with dynamic objects. In this paper, a compact semantic SLAM framework is proposed, with utilization of both geometric and object-level semantic constraints jointly, a more consistent mapping result, and more accurate pose estimation can be obtained. Two main contributions are presented int the paper, a) a robust and efficient SLAM data association and optimization framework is proposed, it models both discrete semantic labeling and continuous pose. b) a compact map representation, combining 2D Lidar map with object detection is presented. Experiments on public indoor datasets, TUM-RGBD, ICL-NUIM, and our own collected datasets prove the improving of SLAM robustness and accuracy compared to other popular SLAM systems, meanwhile a map maintenance efficiency can be achieved.     
\end{abstract}


\section{Introduction}

 \lettrine{I}{n} robotic community, Lidar is a powerful tool for SLAM, but its association relying only on point and line features increases the difficulty of matching and relocalization in dynamic world, even a scene with the lack of appearance, not to mention that further high-level semantic task can be performed on this occupancy map. Though the 2D occupancy can be memory-saving for large scale mapping \cite{carto}, the utilization of such representation remains troublesome.

To overcome the above limitations, exploration of semantic and geometric information fusion, tries to integrate object recognition to carry out object-level mapping, as \cite{SLAM++, kinectfusion++, kinectfusion}. The state-of-the-art neural network for object recognition showcases its ability to handle multi-object classification simultaneously. Acquiring a real-time detection performance is vital for real robotic applications. Many popular structures are proposed, like "R-CNN" \cite{fast-rcnn}  and its variants \cite{faster-rcnn}, by using selective search, the frame rate can speed up, but it requires a lot of work at training phase. Recent work tries to extend the "Deep Learning" to perform 3D perception directly on 3D points, as in \cite{Pointnet, Pointnet++}. Furthermore, \cite{mask-rcnn} adopts object 2D mask to bound the 3D search region as in \cite{frustum-net}, but their real-time performance is still far from practical case. As "Yolov3" \cite{yolo} has evolved into the 3rd generation, a relative high detection speed can be easily realized, compared to other types of detection algorithm.
 
The ability of scene understanding increases the storing map's property. A multi-hierarchical graph \cite{multi-hierarchy}, establishes the connection between spatial and semantic information. These connections, also referred to as anchoring \cite{anchoring}.  Further clarification of hierarchy of semantic abstraction can be found in \cite{multi-hierarchy mobile}.

\begin{figure}[thbp] \centering
   \begin{multicols}{2}
    \begin{subfigure}[thbp]{\linewidth}    
        \includegraphics[width=43mm,height=30.5mm]{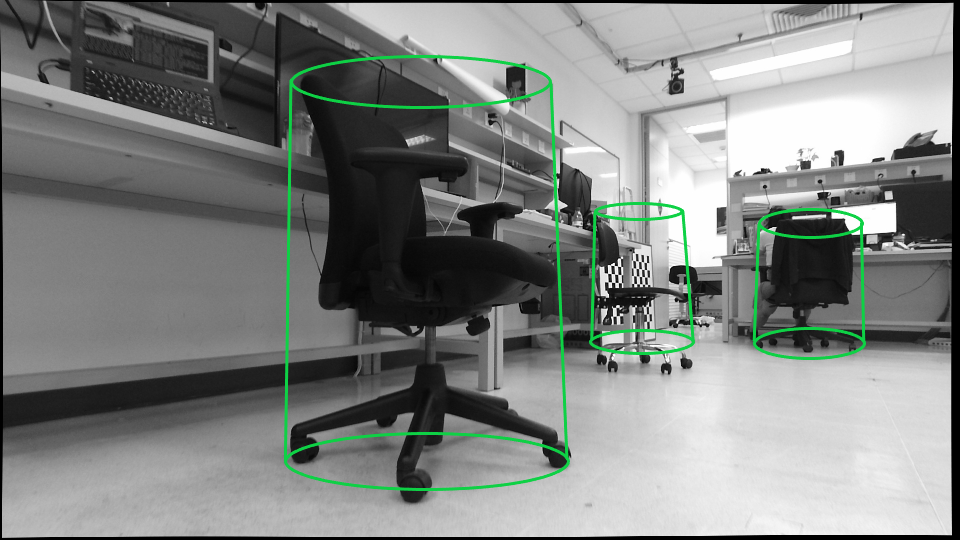}
        \label{fig:cylinder}    
    \end{subfigure} 
     \begin{subfigure}[thbp]{\linewidth}    
        \includegraphics[width=43mm,height=30.5mm]{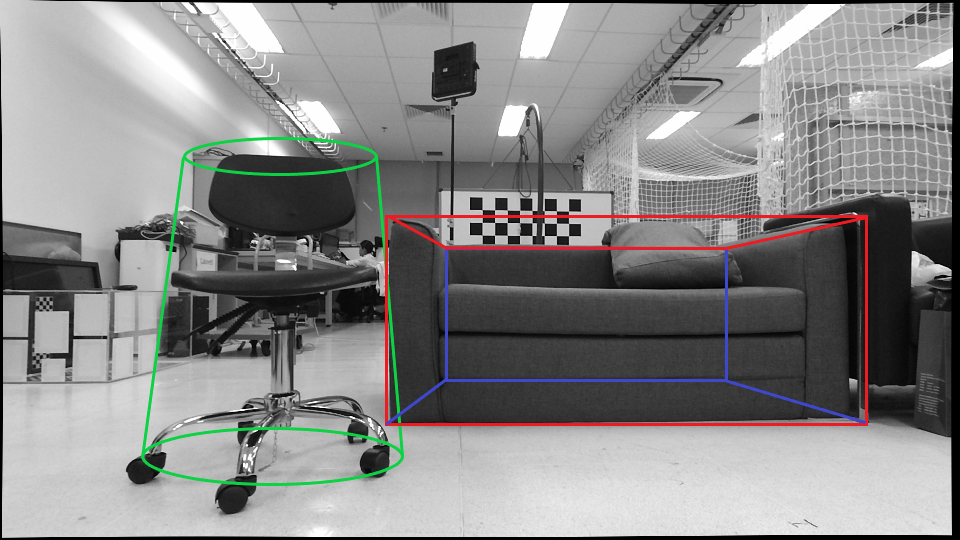}
        \label{fig:cuboid}    
    \end{subfigure} 
  \end{multicols}  
    \centering
    \caption{\small Illustration of our SLAM using cylinder and cuboid as landmark representation. The model scale is predefined. Our 3D model pose estimation in camera frame, mainly depends on observation of visible features, and bottom line of detection bounding box (bbx).}
    \label{fig:3D-2D-model}
\end{figure}

Our first contribution here is the proposal of a robust detector, to generate 3D parameterized landmark from 2D detection. A 2D fast detector depending on RGB-D images, captures both geometric and semantic information, it combines the detection output from convolutional neural network, with geometrical element, such as, points, lines, and simplified parametric model. Followed by a Conditional Random Field (CRF) \cite{CRF} processing the 3D proposals, to estimate the landmark consistently. Secondly, the specified SLAM optimization, over both continuous and discrete variables is introduced, along with how to solve the association problem, with help of object label.  

The paper is structured into six parts, the overall design of system is given at first, then object detector, 3D proposal filtering, and SLAM optimization are explained. Experimental results on public datasets and our own are presented. Finally, a conclusion as summary is made at the end of this paper.

\section{Related Work}

In vision-aiding system, image histogram or scale-invariant feature descriptor search is applied to find match candidate, while the wrong decision leads to catastrophic failure when outliers occur. \cite{vision HMM, reco HMM} selects the most likely appearing place of image probabilistically, via HMM (Hidden Markov Model). HMM can also be applied to Lidar Mapping for obstacle avoidance, discerning navigable area from non-navigable area \cite{auto HMM}.

Data association can be modeled in a probabilistic form, FastSLAM family, \cite{SLAM with unknown DA, FastSLAM, FastSLAM2}, addresses the unknown measurement association incrementally. \cite{Improving-DA} adopts Rao-Blackwellized particle filter to implement data association. \cite{Connecting-Dominating-Sets} clusters a subset of images to represent all previously seen images, so that a constant updating time complexity can be assured as map grows. 

Extended Kalman Filter (EKF) based SLAM can combine the object-level landmark, as work done in \cite{DA-EKF-SLAM}, but all landmarks are limited only to planar models, so a homography transform from scene to image should be estimated. For robot moving on ground, EKF can generate a map with 3D landmarks \cite{Ceiling-SLAM}. \cite{association} proposes an EM based approach to optimize discrete semantic label, along with continuous geometric variable, jointly in a probabilistic way, but from their demo video, the landmark seems to be initialized prior to SLAM.

The real-time performance is our main concern, so method with simplifying model \cite{Quadric-Slam, Plane-Slam, Cube-Slam}, quadrics, cuboids, can represent object position and orientation in a compact way, but still helpful to robot navigation. 

\section{Overall Structure}

The system overview is shown in Figure \ref{fig:outline}, to be noted, that box in pink is integrated from open-source code. Green box implies our main code implementation. The whole software runs on Ros platform. The calibration for camera intrinsic parameters, cross calibration for rgb and depth frame, extrinsics between camera and 2D laser is done beforehand. Especially, cross calibration of Kinect and Lidar is done via controlling point, co-observed by both sensors, and associated manually, finally iterative optimization with "ceres" tool \cite{ceres} can provide a feasible extrinsic configuration. The whole transform tree is maintained by Ros "tf". The detector still works on 2D image, and database now is only a simplified version with swivel chair, door, Sofa, with known dimension. Content of detector is expanded in Figure \ref{fig:detector}. These selected furniture is quite common in an indoor scene, they can be easily scaled and re-designed, and the assumption that they are in ground plane can hold in reality.

\pgfdeclarelayer{background}
\pgfdeclarelayer{foreground}
\pgfsetlayers{background,main,foreground}
\tikzstyle{blockIn} = [draw, fill=blue!10, text width=5em, 
    text centered, minimum height=2em]
\tikzstyle{blockOut} = [draw, fill=white, text width=5em, 
    text centered, minimum height=2em]
\tikzstyle{blockFunc} = [draw, fill=red!10, text width=7em, 
    text centered, minimum height=5em,drop shadow]
\tikzstyle{blockDA} = [draw, fill=green!10,text width=5em, 
    text centered, minimum height=14em,drop shadow]
\tikzstyle{cloudDB} = [draw, ellipse,fill=blue!30, text width=7em,text centered, minimum height=2.5em, drop shadow] 

\begin{figure}[thbp]
  \centering
  \resizebox{.5\textwidth}{!}{
  \begin{tikzpicture}
  \node (rgb) [blockIn] {RGB};
  \path (rgb.east)+(2.5,+0.0) node (objdet) [blockFunc] {Object Detector (2D$\,\to\,$3D)}; 
  \path (rgb.south)+(0.0,-2.0) node (depth) [blockIn] {Depth}; 
  \path (depth.east)+(2.5,0.0) node (orb) [blockFunc] {ORB2 Tracker}; 
  \draw [->] (rgb) -- (objdet);
  \draw [->] (depth) -- (objdet);
  \draw [->] (rgb) -- (orb);
  \draw [->] (depth) -- (orb);
  \path (objdet.east)+(2.0,0) node (objloc) [blockOut] {Object Location (cam. coord.)};
  \path (orb.east)+(2.0,0) node (campose) [blockOut] {Camera Pose};
  \draw [->] (objdet) -- (objloc);
  \draw [->] (orb) -- (campose);
  \path (objdet.north)+(0,+1.1) node (objdb) [cloudDB] {Object Database (label + 3D size)};
  \draw [->] (objdb) -- (objdet);
  \path (depth.south)+(0.0,-2.5) node (lidar) [blockIn] {Lidar};
  \path (lidar.east)+(2.5,0) node (dynobj) [blockFunc] {Dynamic Object Filter};
  \path (dynobj.east)+(2.3,0) node (cart) [blockFunc] {Cartographer};
  \draw [->] (lidar) -- (dynobj);
  \draw [->] (dynobj) -- (cart);
  \path (campose.east)+(2.0,0.3) node (da) [blockDA] {Data Association};
  \draw [->] (objloc) -- (da);
  \draw [->] (campose) -- (da);
  \draw [->] (dynobj) -- (da);
  \path (da.east)+(1.5,0) node (landmark) [blockOut] {3D Semantic Landmark};
  \path (cart.east)+(3.7,0) node (gridmap) [blockOut] {2D Occupancy Grid Map}; 
  \draw [->] (da) -- (landmark);
  \draw [->] (cart) -- (gridmap);
  \path (landmark.east)+(1.8,-1.5) node (insertion) [blockFunc] {Landmark Insertion};
  \path (insertion.north)+(0,2.9) node (tf) [blockFunc] {TF};
  \draw [->] (landmark) -- (insertion);
  \draw [->] (gridmap) -- (insertion);
  \path (insertion.east)+(1.5,0) node (map) [blockIn] {2.5D Semantic Map};
  \draw [->] (insertion) -- (map);
  \draw [->] (tf) -- (insertion);
  \draw [->] (tf.west) -- (da);
  \begin{pgfonlayer}{background}
    \path (objdb.west |- objdb.north)+(-0.3,0.4) node (A) {};
    \path (campose.east |- campose.south)+(+0.3,-0.8) node (B) {};
    \path[fill=yellow!20,rounded corners, draw=black!50, dashed] (A) rectangle (B);                    
  \end{pgfonlayer}
  \end{tikzpicture}}
  \caption{\small{System Framework, box in white is data output, box in pink or green is the separate thread for specifying task. Top left with yellow background is visual front-end, extracting features from RGB-D image pair, aligning temporally and spatially in camera frame. The detector performs 2D geometrical element extracting, and semantic labelling in real-time for each frame. Model parameter predefined in database will be loaded according to detection output, generating 3D object pose estimation in camera. 2D Lidar scan points passing through the filter will be separated into static or dynamic parts, then points belonging to static object, along with output from camera block, will be utilized as input of data association, to provide a consistent landmark association result. Top right is a "tf" tree buffering the transform between different frames. Bottom right corresponding to static part in each scan, will be fed into "cartographer" \cite{carto}, to generate 2D occupancy map, and new created landmark with object id will be inserted onto occupancy map.}}
  \label{fig:outline}
\end{figure}
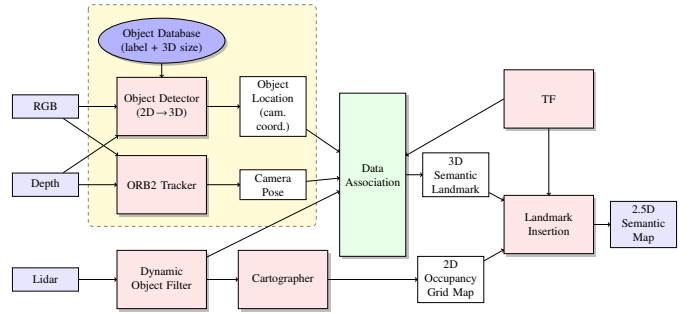  

Following part will make a detailed clarification on core design.

\section{2D-3D detector}\label{Sec IV}

The pre-trained model on "COCO" \cite{COCO} can detect above 80 classes, then we further fine tuned the network on "ICL-NUIM" \cite{NUIM} and "SUN-RGBD" \cite{SUN-RGBD}, to strengthen the network inference performance in indoor scene. We also collect some samples for door handle in our work-place building to improve successful rate of door detection. After training, the network can classify chair, Sofa, person, and door with AP above 65\% in general case. 

As in Figure \ref{fig:line}, the long Canny edge, detection box size, along with object depth mask, which is generated from point-cloud clustering after ground plane removal, will score for the initialization condition. The bigger box size, more visible edges, more valid per pixel depths are present, a higher weight value on landmark pose will be applied for SLAM post-optimization. 

\begin{figure}
    \centering
        \includegraphics[scale = 0.25]{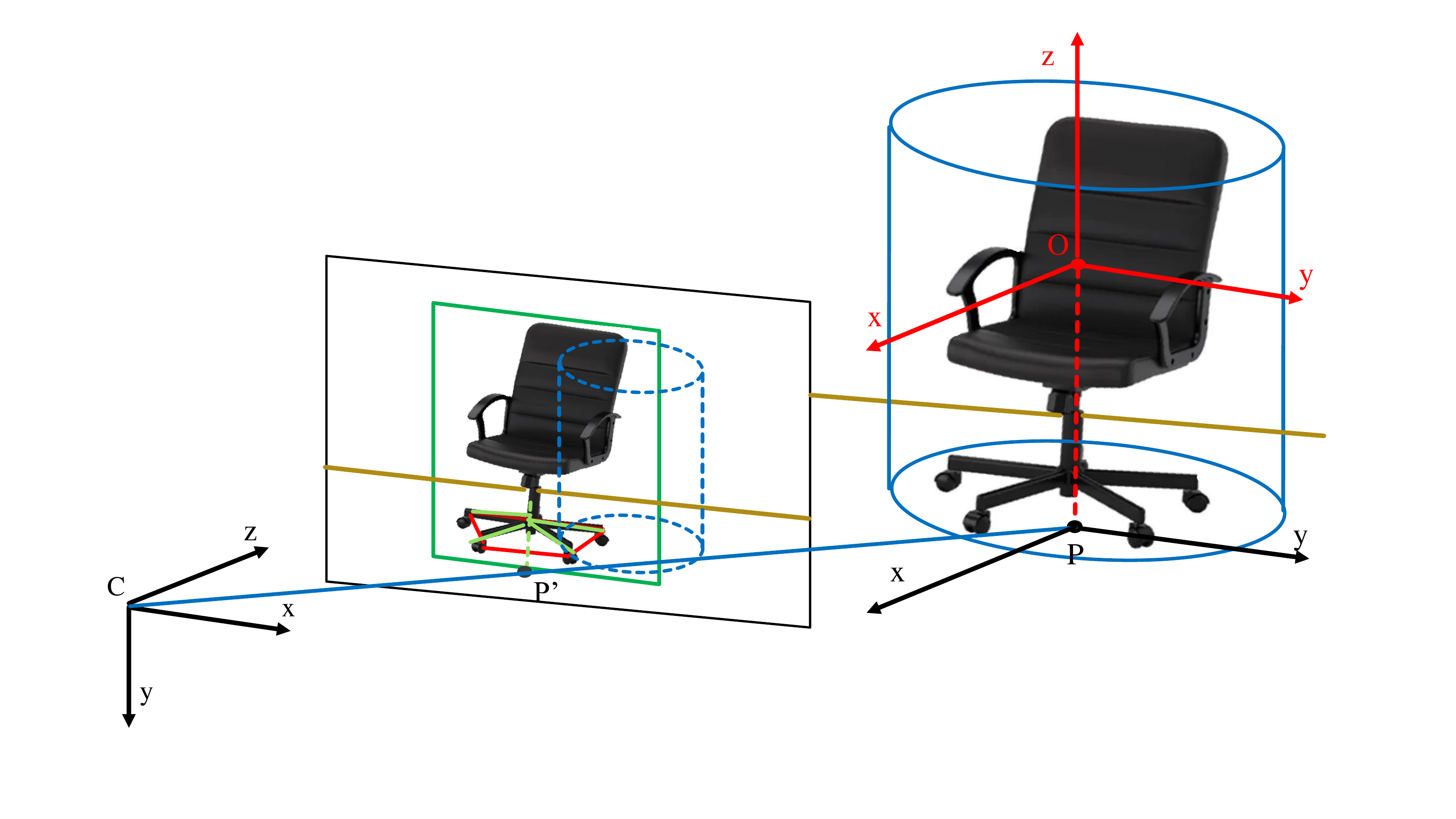}     
        \caption{Back Projection of Corner Pixel in Ground Plane}
    \label{fig:my_pdf}
\end{figure}

The fundamental concepts around view geometry are presented as Figure 2. \\
A. Detection sensor model for swivel chair. The bottom part of a revolving chair has a single leg, the corresponding detected line segment can be extended to intersect the bottom line of bbx at $P'$. This point is supposed to be in ground plane, a ray from camera center to this pixel meets the ground plane at a point, the object centroid is with a fixed above height. When the line detection is ill-conditioned, the bottom edges near bottom line, via Hough transform can vote for multiple intersection points, further extended along $-y$ direction to provide multi-hypothesis for corner in ground, these grounding point proposal will be opted only when its IOU is above 0.3. IOU is calculated through re-projection model, between dashed cylinder in blue, and bounding box in green on image plane, as shown in Figure 3. 

Cylinder Height $h$ and radius $r$ are given. From the ground plane model removal process above, the normal $\vv{n}$ of ground point cloud in camera frame can be estimated, then this plane is transformed from world to camera frame with extrinsic transform $\vv{P} = [R, \vv{t}]$, intrinsic parameters are encoded in $K$. The difference error between transformed normal vector and initial normal vector can also be used as SLAM optimization constraint. The camera center $\vv{{C}_c} = [0, 0, 0]$ in camera frame, substitute it into left side of Equation 1, the center can be expressed in ground plane frame, $\vv{{C}_g} = -{R}^T * \vv{t}$. The homogeneous pixel coordinate in ground plane is $\vv{m} = [u, v, 1]$, in camera frame it's $\vv{{m}_c} = {K}^{-1} * \vv{m}$, plug it into Equation 1, $\vv{{m}_g} = {R}^T * (\vv{{m}_c} - \vv{t}) $. Equation 2 is the ray function from camera center $\vv{{C}_g}$ through the corner pixel $\vv{{m}_g}$. $\vv{{X}_g} = ({x}_g, {y}_g, {z}_g)$ is a point in ground plane frame, ${z}_g$ is known to be $h/2$. So three unknowns, ${x}_g$, ${y}_g$, $\lambda$, can be solved by three equations. The cylinder center is $({c}_x, {c}_y, {c}_z) = ({x}_g, {y}_g, h/2)$. Finally the top and bottom circle centers can also be determined.

The ground normal vector in camera frame will be estimated along with the cylinder proposal, because the 3D proposal is estimated in a local ground plane of camera, the projected model onto image may have a gentle pitch or roll phenomenon, this inconsistency can be reflected by size inconsistency of top and bottom projected circles of cylinder, as in Figure \ref{fig:my_pdf}.

\begin{align}
        &\vv{{X}_c} = R * \vv{{X}_g} + \vv{t}\\
        &\vv{{X}_g} = \vv{{C}_g} + \lambda * (\vv{{m}_g} - \vv{{C}_g})
\end{align}

B. Detection sensor model for Sofa. As for Sofa, the visible line should be more obvious, to form the box edges. Previous work by \cite{Cube-Slam} is introduced here, a common sampling over orientation to get vanishing point as initial step, followed by intersecting 2D corner from vanishing points and top corner point, the 2D corners defined the cuboid location. The best sampling proposal will be chosen according to a complex scoring process. Multiple visible long edges along Sofa can be some clue for proposal selection.

\begin{figure}[thbp] \centering
   \begin{multicols}{2}
    \begin{subfigure}[thbp]{\linewidth}    
        \includegraphics[width=43mm,height=28.5mm]{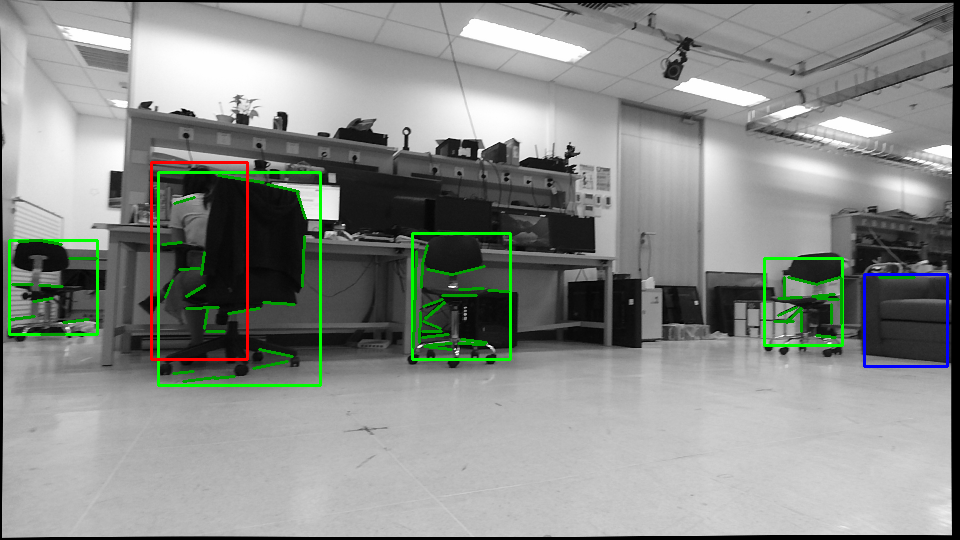}
        \label{fig:line-b}    
    \end{subfigure} 
     \begin{subfigure}[thbp]{\linewidth}    
        \includegraphics[width=43mm,height=28.5mm]{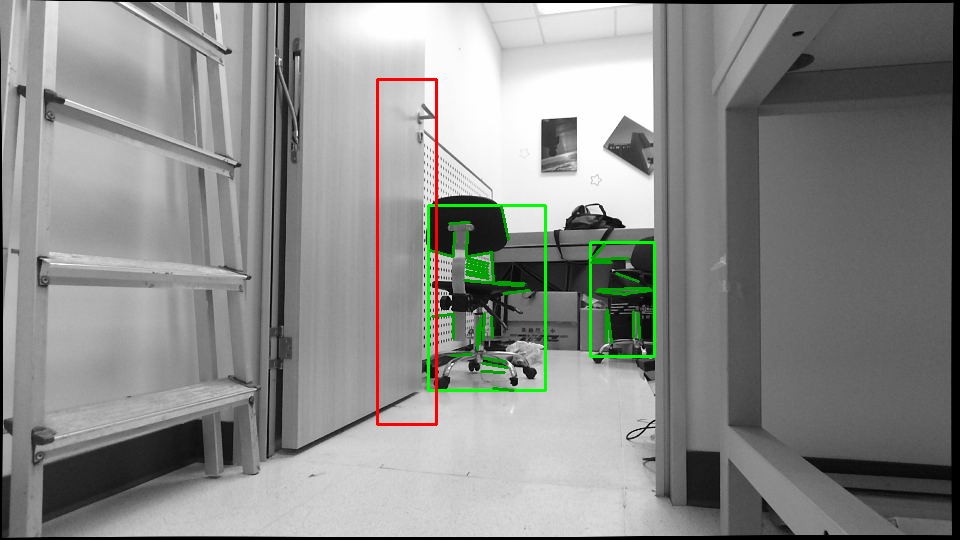}
        \label{fig:line-c}    
    \end{subfigure} 
  \end{multicols}  
    \centering
    \caption{\small Line Detection within Semantic Bounding Box}
    \label{fig:line}
\end{figure}

\begin{figure}[thbp]   
   \begin{multicols}{2}
    \begin{subfigure}[thbp]{\linewidth}    
        \includegraphics[width=43mm,height=28.5mm]{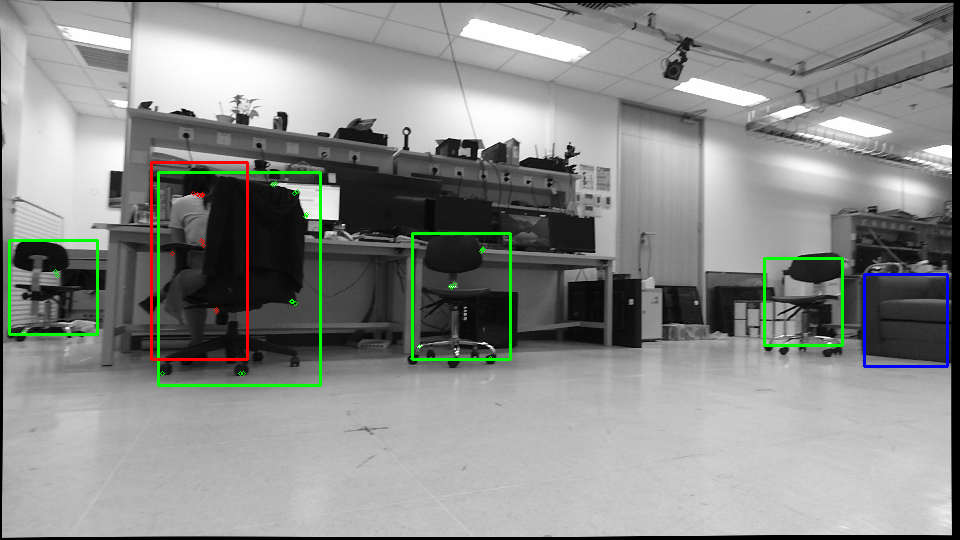}
        \label{fig:point-b}    
    \end{subfigure} 
    \begin{subfigure}[thbp]{\linewidth}    
        \includegraphics[width=43mm,height=28.5mm]{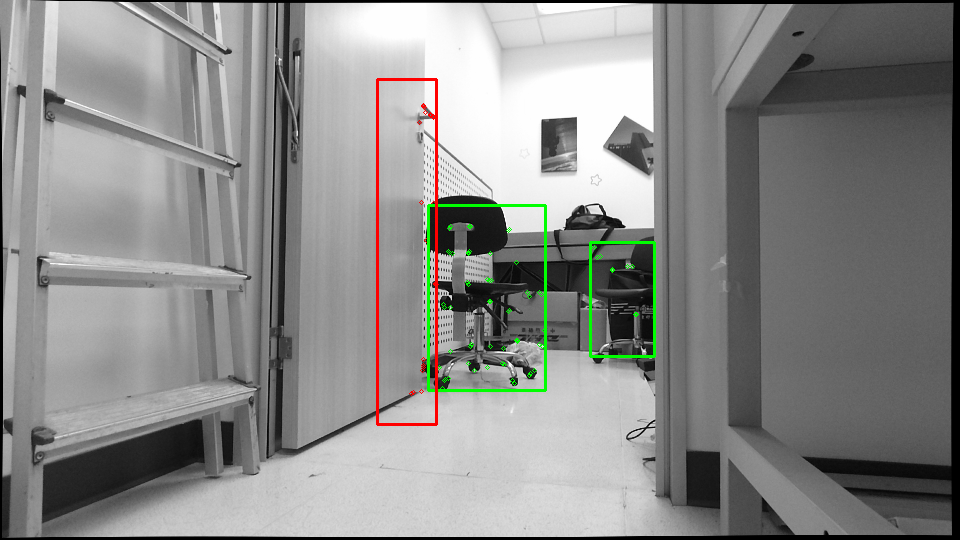}
        \label{fig:point-c}    
    \end{subfigure} 
  \end{multicols}  
    \centering
    \caption{\small Tracking ORB Features within Semantic Bounding Box}
    \label{fig:point}
\end{figure}

\tikzstyle{decision} = [diamond, draw,fill=green!10, text width=6em,minimum height=5em,drop shadow]
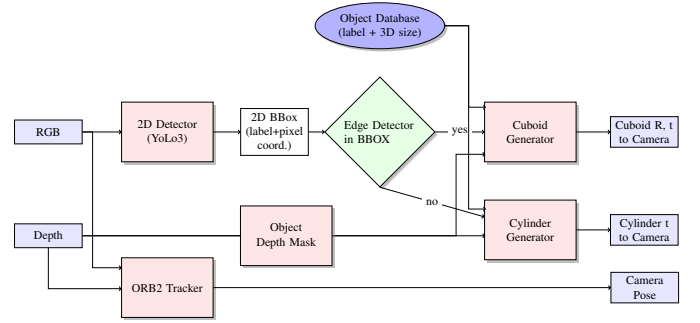
\begin{figure}[H] \centering
  \resizebox{.5\textwidth}{!}{
  \begin{tikzpicture}
  \node (rgb) [blockIn] {RGB};
  \path (rgb.east)+(2.5,+0.0) node (yolo) [blockFunc] {2D Detector (YoLo3)}; 
  \path (yolo.east)+(1.8,0) node (bbox) [blockOut] {2D BBox (label+pixel coord.)};
  \path (bbox.east)+(2.1,0) node (edge) [decision] {Edge Detector in BBOX};
  \path (edge.east)+(2.8,0) node (cube) [blockFunc] {Cuboid Generator};
  \path (cube.east)+(2,0) node (cuboid) [blockIn] {Cuboid R, t to Camera};
  \path (edge.north)+(0,1.5) node (objdb) [cloudDB] {Object Database (label + 3D size)};
  \draw [->] (rgb) -- (yolo);
  \draw [->] (yolo) -- (bbox);
  \draw [->] (bbox) -- (edge);
  \path [draw, ->] (edge) -- node [midway,fill=white] {yes} (cube);
  \draw [->] (cube) -- (cuboid);
  \path (cube.south)+(0,-2) node (cylinder) [blockFunc] {Cylinder Generator};
  \path (cylinder.east)+(2,0) node (cyl) [blockIn] {Cylinder t to Camera};
  \path (rgb.south)+(0,-2.7) node (depth) [blockIn] {Depth};
  \path (yolo.south)+(0,-3.7) node (orb) [blockFunc] {ORB2 Tracker};
  \path (orb.east)+(12.7,0) node (cam) [blockIn] {Camera Pose};
  \draw [->] (orb) -- (cam);
  \draw [->] (cylinder) -- (cyl);
  \path [draw, ->] (edge.south) -- node [midway,fill=white] {no} (cylinder);
  \draw[->] (depth.south) -- ++(0.0,-1.2) -- ++(+2.15,0.0) node {} (orb);
  \draw[->] (rgb.east) -- ++(0.3,0) -- ++(+0,-4) --++(0.85,0) node {} (orb);
  \draw[->] (depth.east) -- ++(11,0) -- ++(+0,2.4) --++(0.85,0) node {} (cube);
  \draw[->] (depth.east) -- ++(11.85,0) node {} (cylinder);
  \draw[->] (objdb.east) -- ++(0.7,0) -- ++(+0,-2.4) --++(0.5,0) node {} (cube);
  \draw[->] (objdb.east) -- ++(0.7,0) -- ++(+0,-5.4) --++(0.5,0) node {} (cylinder);
  \path (depth.east)+(6,0) node (edgfilter) [blockFunc] {Object Depth Mask};
  \end{tikzpicture}}  
  \caption{\small Detector Diagram, the whole detection process is performed over the rgb-d image pair, with estimated cylinder, indicating swivel chairs, or cuboid for Sofa, the person is considered as outlier, the door detection will be a reminder for new landmark initialization. Feature points and long edges, only within detection bounding box are considered. Directly ORB feature tracking \cite{ORB} will provide relative transform between consecutive frame poses. The depth mask is used only to verify the valid edges within object mask.}
  \label{fig:detector}
\end{figure}

\section{2D-3D Proposal Selection}\label{Sec V}

In the following sections, the proposal selection method and SLAM optimization will be formulated in mathematical representations. 

\subsection{Proposal Generation}
As the detection part explained in Section \ref{Sec IV} B, on-the-fly detection will generate multiple proposals in each frame, though the harsh verification condition can filter out a lot of spurious detection candidates, to get the best proposal a further selection should be gone through.

\subsection{CRF Modeling}\label{crf}
Here our CRF model, Equation \ref{eq:3} suggests a multi-level potential. 
\begin{equation}
\begin{split}
    E(\bold{x}|\bold{b}) &= \sum_{j}\sum_{i} {\alpha}_{j} \Psi^{U}({x}_{ij}^{t}) + s(1 - \beta)\Psi^{P}_{\beta}({\bold{x}}_{}^{t-1}, \bold{x}_{}^{t}) \\ &+ \sum_{{x}_{*j}^{n} \in \bold{C}} (1 - {\beta}^{*})\Psi^{{H}}({x}_{ij}^{n})
\end{split}
\label{eq:3}
\end{equation}

1) \textit{Unary potential}
The unary energy implies the quality of object proposals. ${x}_{ij}^{t} \in [0 , 1]$, indicating the $i$th object proposal belonging to the object $j$, in image frame at $t$, ${\alpha}_{j}$ is the classification probability. For proposal of same object id, it is constant.

\begin{equation}
\begin{split}
    \Psi^{U}({x}_{ij}^{t}) &= -s({x}_{ij}^{t})(1 - d({c}_{{x}_{ij}^{t}}, {c}_{bbx}))
\end{split}
\label{eq:4}
\end{equation}

The overall negative sign on the right side of equation indicates that there is an encouraging for more proposals. $s({x}_{ij}^{t})$ is the ratio of projected 3D model size over image size, a large ratio indicates a small error. $d({c}_{{x}_{ij}^{t}}, {c}_{bbx})$ is a normalized distance between detection box centroid, and 3D proposal centroid on image, indicating that the larger this distance is, the more penalty comes in. 

2) \textit{Pairwise potential}
This part is to model the semantic association between two consecutive frames. $s$ is a normalized ratio from 0 to 1, reflecting how many shared feature points are in bbx belonging to the same instance. $\beta$ implies a matching score for semantic information. The semantic label appearing in each frame will be encoded into a binary sequence, from top left of image to bottom right. ${b}^{t-1}$, ${b}^{t}$ indicate the encoding semantic sequence of last frame, and current frame respectively.

\begin{equation}
\begin{split}
    \beta &= \frac{l({b}^{t-1} \cap {b}^{t})} {l({b}^{t-1} \cup {b}^{t})}
\end{split}
\label{eq:5}
\end{equation}

The more matching pattern bits there are in two frames, the higher value will be scored. $l$ represents the length size operation.

\begin{small}
\begin{equation}
\begin{split}
    \Psi^{P}_{\beta}({\bold{x}}_{}^{t-1}, \bold{x}_{}^{t}) &= {x}_{ij}^{t} {x}_{mn}^{t-1}\sum_{\gamma(j)}\sum_{m}\sum_{j}\sum_{i} (1-\frac{A(x_{ij}^{t}) \cap {H}_{t-1}^{t}A(x_{mn}^{t-1})} {A(x_{ij}^{t}) \cup {H}_{t-1}^{t}A(x_{mn}^{t-1})})
\end{split}
\label{eq:6}
\end{equation}
\end{small}

The pairwise potential has a rather perplexing form, when co-occurrence happens, the matching error will be considered, Here ${H}_{t-1}^{t} = P_{t}P_{t-1}^{-1}$, the $P$ indicates the projective transform, more overlapping area between predicted proposal area and observed proposal area, should be encouraged. $\gamma(j) \in n$, is a mapping, through which we can get the queried object id index on the other frame, this step is done by binary recursive search, from longest matching pattern to the smallest, with return of the ordering number, it can establish all possible association between the same kind of object's proposals in different frames. In spite of four sum loops, the iteration over object $j$ and $\gamma(j)$ can be rather small with limited counts, so time complexity is approximately $O(n^2)$.

3) \textit{High order potential}
For high order potential, at most, only one of the 3D object proposals per landmark id will be selected. The clique $C$ can be a subset of instances, corresponding to a local window over multiple frames, that are very close both temporally and spatially. The clique only includes object instance inside this window.

\begin{equation}
\Psi^{{H}}({x}_{ij}^{n}) = \begin{cases}
0 ,& if\quad 0\leq\sum\limits_{n = 1}^{N}{\sum\limits_{i}}{x_{ij}^{n}} \leq N\\
\infinity & otherwise
\end{cases}
\label{eq:7}
\end{equation}{}

In third term of Equation \ref{eq:3}, ${\beta}^{*} = \frac{l({b}^{t} \cap {b}^{t-k,...,t})} {l({b}^{t} \cup {b}^{t-k,...,t})}$, it is length of binary matching bits similar to the that in \textit{Pairwise potential}, the difference is the binary pattern is against a sequence, encoding all semantic information within local window. Fast binary search algorithm \cite{binary-string-search} is still quite challenging. $N$ states the number of frames in dynamic window. A good proposal should appear at most once, but at least once in the $N$ frames.

\section{SLAM Optimization}

From above CRF model, a most reasonable configuration of proposals is determined, that should be consistent with observation up-to-now. A joint distribution over landmark and robot pose can be derived, composed of landmark pose $L=\{{l}_{j}\}$ and its binary semantic label $C=\{{c}_{j}\}$, feature point in bbx $P=\{{p}_{n}\}$, detection model $B=\{b_{ij}\}$, denoting the detection probability of object $j$ observed in frame $i$, $X=\{x_i\}$ indicates the robot pose, measurement association $D={\beta_{kj}}$, assigning a measurement $k$ to landmark $j$. The measurement includes both contiguous and discrete variables, thence the problem can be formularized as following, 

\begin{equation}
\begin{split}
\textit{p}(L, P, X|U, B, D) &= \eta\bold{\prod_{i}} \textit{p}(x_{i+1}|x_{i}, u_{i}) \bold{\prod_{n,i, j}}\textit{p}(p_{n}|{l_{j}, x_{i}, \beta_{nj}})\\
& \quad \bold{\prod_{i,j,k}}\textit{p}(l_{j}|x_{i},b_{ij},\beta_{kj})\bold{\prod_{n,i,j}}\textit{p}(p_{n}|x_{i}, l_{j}, b_{ij})
\label{eq:8}
\end{split}{}
\end{equation}

$\eta$ in Equation \ref{eq:8} is a normalizer, After taking a negative log over the equation two sides, the four factorized terms above, can be simplified into a summed form, equivalent to solving for minimal squared Mahalanobis distance as least square problem.

\begin{small}
\begin{equation}
\begin{split}
X*, P*, L* &= {arg\,min}\,-log(\textit{p}(L, P, X|U, B, D))\\ &={arg\,min}\,\sum_{i}||\bold{e}(x_{i+1}, f(x_{i}, u_{i}))||^{2}_{\sum_{i}} +\\& \sum_{n}||\bold{e}(\uppi(p_{n},x_{i}),\uppi(p_{n},x_{i+1}))||^{2}_{\sum_{ni}} + \sum_{j}||\bold{e}(l_{j}, g(b_{ij}, x_{i}))||^{2}_{\sum_{ij}} +\\& \sum_{n}||\bold{e}(p_{n}, g(b_{ij}, x_{i}))||^{2}_{\sum_{ij}}
\label{eq:9}
\end{split}{}
\end{equation}
\end{small}{}

The above factors indicate odometry factor, camera to point factor, camera to landmark factor, and point to object factor, respectively, given the measurement association is solved by equation below.

\begin{equation}
\begin{split}
\beta_{kj}^{*} &= arg\,max\,\,log\,\textit{p}(\beta_{kj}|b_{i*}, p_{n}, x_{i}, c{*})\\ &= arg\,max\,\,\omega_{kc}\prod_{t}\textit{p}^t_{0}\textit{p}^t_{c} \sum_{j}\frac{\textit{p}(b_{ij}|x_{i},l_{j},\beta_{kj}^{t})}{\sum\limits_{j}\textit{p}(b_{ij}|x_{i},l_{j},\beta_{kj}^{t})}
\label{eq:10}
\end{split}
\end{equation}

At every step, the coordinate descent over Equation \ref{eq:9}, \ref{eq:10} will be iterated, so that point or 3D model proposal as measurement $k$ will be assigned to the proper landmark. From aforementioned part, each detection instance on image to all possible landmarks probability will be summed, the $p_{0}^{t}$ as a normalized prior denotes the tracking feature points' number in bbx occuring at frame $t$, ${p}^t_{c}$ indicates the semantic prior for frame $t$, proportional to binary matching length $\beta$ as clarified in Section CRF. To every measurement. $\omega_{kc}$ is a binary indicator, only as one when the measurement label comes in accordance with the landmark $j$'s label,  measurement here represents the detected bounding box in each frame, for each detected instance this sum-product propagation will be performed recursively, to compute the association distribution approximately. $p(b_{ij}|x_{i},l_{j},\beta_{kj}^{t})$ is a distribution, over the euclidean distance of measurement in 3D to the landmark centroid. When the whole term drops below a threshold, a new landmark should be created.

1)Pose to pose constraint: The camera pose is extrapolated from EKF output temporally, a constant velocity model is assumed for tracking, the difference between prediction and observation is calculated on $SE(3)$ space.

2)Camera to point constraint:  
This is a normal implementation of 3D point re-projection error in \cite{ORB},  
\begin{equation}\label{eq:11}
e_{cp} = (\uppi(p_{n}x_{i}^{-1}) - p') + (\uppi(p_{n}x_{i+1}^{-1}) - p'')
\end{equation}
$p_{n}$ is in world frame and firstly transformed to camera frame, followed by 2D projection, $p'$ and $p''$ are observed pixel coordinates in image.

3)Camera to landmark constraint: 
The 3D proposal should be transformed to the camera frame first, then a bounding box enveloping the projected shape can be formed by find four corners. An exception case is when only a portion of projected model within image, the approach "adjugate" \cite{Quadric-Slam} can be applied to compute the bounding box corner aligned with the main axis of ellipse.

\begin{align}
\label{eq:12}
e_{cl} &= \hat{b}_{ij} - b_{ij}\\
\hat{b}_{ij} &= min/max\,{\uppi(l_{j}x_{i}^{-1})}
\end{align}
$b_{ij}$ is a 4D vector including top-left and bottom-right corner points' coordinate.

4)Point to landmark constraint:
A truncation is used to make sure the 3D feature point fitting into the 3D model dimension, the operation is done by 2D Euclidean distance plus one dimension range check along $z$ axis.
\begin{equation}\label{eq:13}
e_{pl} = max(p_{n}l_{j}^{-1} - c_{l_{j}}, 0)
\end{equation}
$c_{l_{j}}$ is the 3D centroid of landmark $j$.
   \begin{figure}[thbp]
     \centering
      \includegraphics[scale = 0.23]{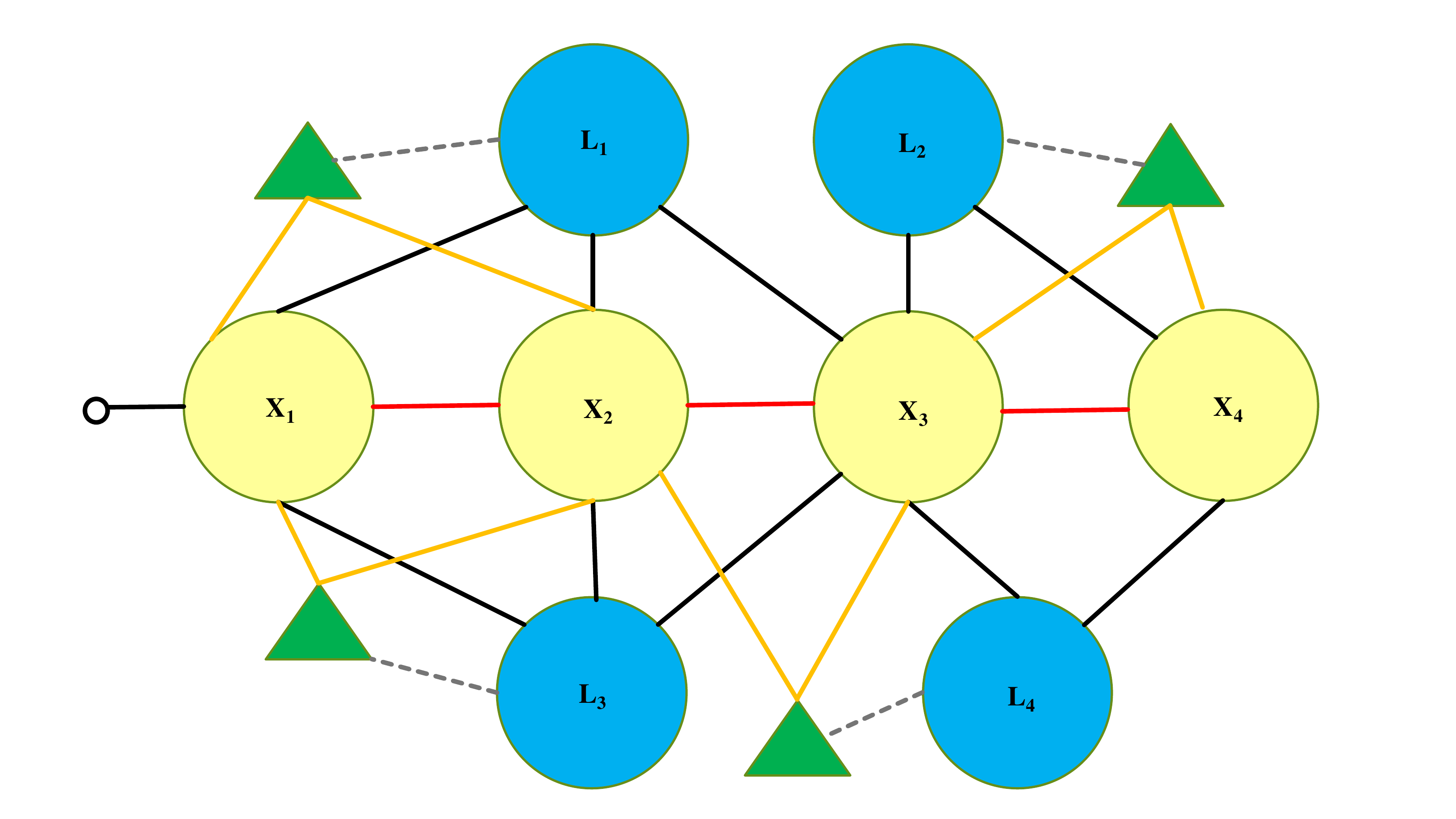} 
      \caption{\small Factor Graph. The camera pose is represented in yellow circle, while blue circle denotes landmark, the small green triangle represents the feature point. The small hollow circle connected to leftmost pose is the 2D ground plane constraint, four types of mutual constraint, point-landmark, pose-pose, pose-point, pose-landmark are denoted by different colored lines, the dashed line indicates that the point to landmark constraint is binary}
      \label{fig:factor }
   \end{figure}  

\section{Experiments}

\begin{figure}[H]
     \centering
      \includegraphics[scale = 0.32]{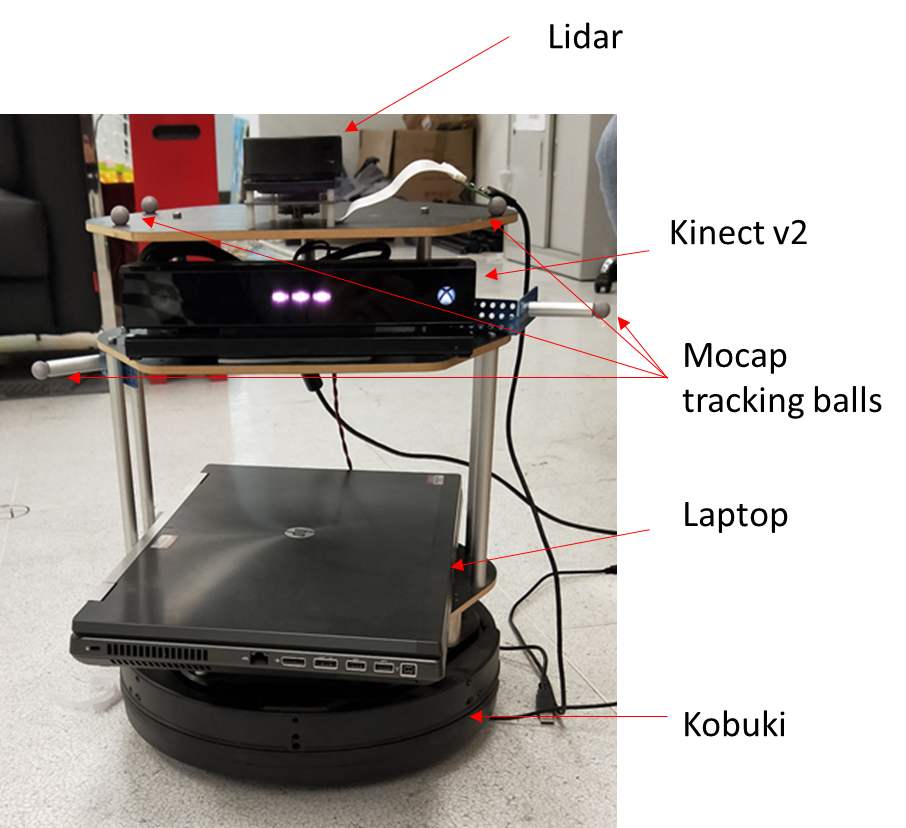} 
      \caption{\small Outfitting Demonstration of Robot Platform.}
      \label{fig:robot}
   \end{figure}

The data sequence is collected on a Kobuki robot, outfitted with "Rplidar" and "Kinect", the tracking balls are also attached onto the rigid body of robot as in Figure \ref{fig:robot}, to produce ground-truth trajectory. All the data are transferred to the "HP Zhan99 G1" laptop, with "Intel i7-8750H \@ 2.2 GHz, NVIDIA Quadro P600 graphical card", all system framework runs on the laptop as a local workstation. The cross calibration between sensor frames were implemented offline. 

\begin{table}[H]
\centering
\footnotesize
\begin{tabular}{l|c|c|c|c}
\toprule
 Approach& ORB2\cite{ORB} w/o LC & Cube+Ours\cite{Cube-Slam} & carto\cite{carto} & ours\\
 \hline
Living1 &0.041 & \bf{0.039} & - & 0.043\\
Living2 & 0.038 & \bf{0.034} & - & 0.045\\
Office1 & \bf{0.062} & 0.073 & - & 0.069\\
Office2 & 0.048 & \bf{0.029} & - & 0.032\\
 \hline
 \hline
fr1/360 & \bf{0.028} & 0.058 & - & 0.062\\
fr1/floor & \bf{0.068} & 1.157 & - & 2.128\\
fr1/desk & \bf{0.021} & 0.028 & - & 0.022\\
fr2/desk & 0.014 & 0.019 & - & \bf{0.013}\\
fr3/office & 0.036 & 0.032 & - & \bf{0.028}\\
\hline
\hline
lab792 & 15.026 & 0.078 & 6.353 & \bf{0.054}\\
\hline
\end{tabular}
\caption{RMSE in ATE (Abosolute Trajectory Error) On Various Datasets(m)}
\label{slam-result}
\end{table}
For comparison, loop closure of ORB-RGBD \cite{ORB} and catographer \cite{carto} are disabled, Cube proposal generation is combined with our association back-end, it's hard to repeat SLAM result \cite{Cube-Slam}, because of its incomplete open-source code missing data association part. Motion prior in 3D is generated from "ORB" \cite{ORB} tracking model, On "ICL-NUIM" the "Cube+Ours" approach outperforms others, because sequences include only Sofa or desktop, even cuboid shaped chair. On "TUM" dataset, our framework can perform well when many required models for detection are present, while in Monotonous scene or with abrupt 3D orientation, the ill-conditioned object detection or wrong ground plane normal estimation can corrupt the whole system as in "fr1/floor". On our self-made dataset "lab792", the robot patrols over between a part of big lab room and a small connected room. Swivel chair and Sofa are distributed along moving trajectory, persons are also present in the room. Our system proves the merits of the constraints introduced at the object level.

The trajectory result on "lab792" generated from different approaches is plotted below.

\begin{figure}[H] 
   \centering
   \begin{multicols}{2}
    \begin{subfigure}[H]{\linewidth}    
        \includegraphics[width=47mm,height=39.5mm]{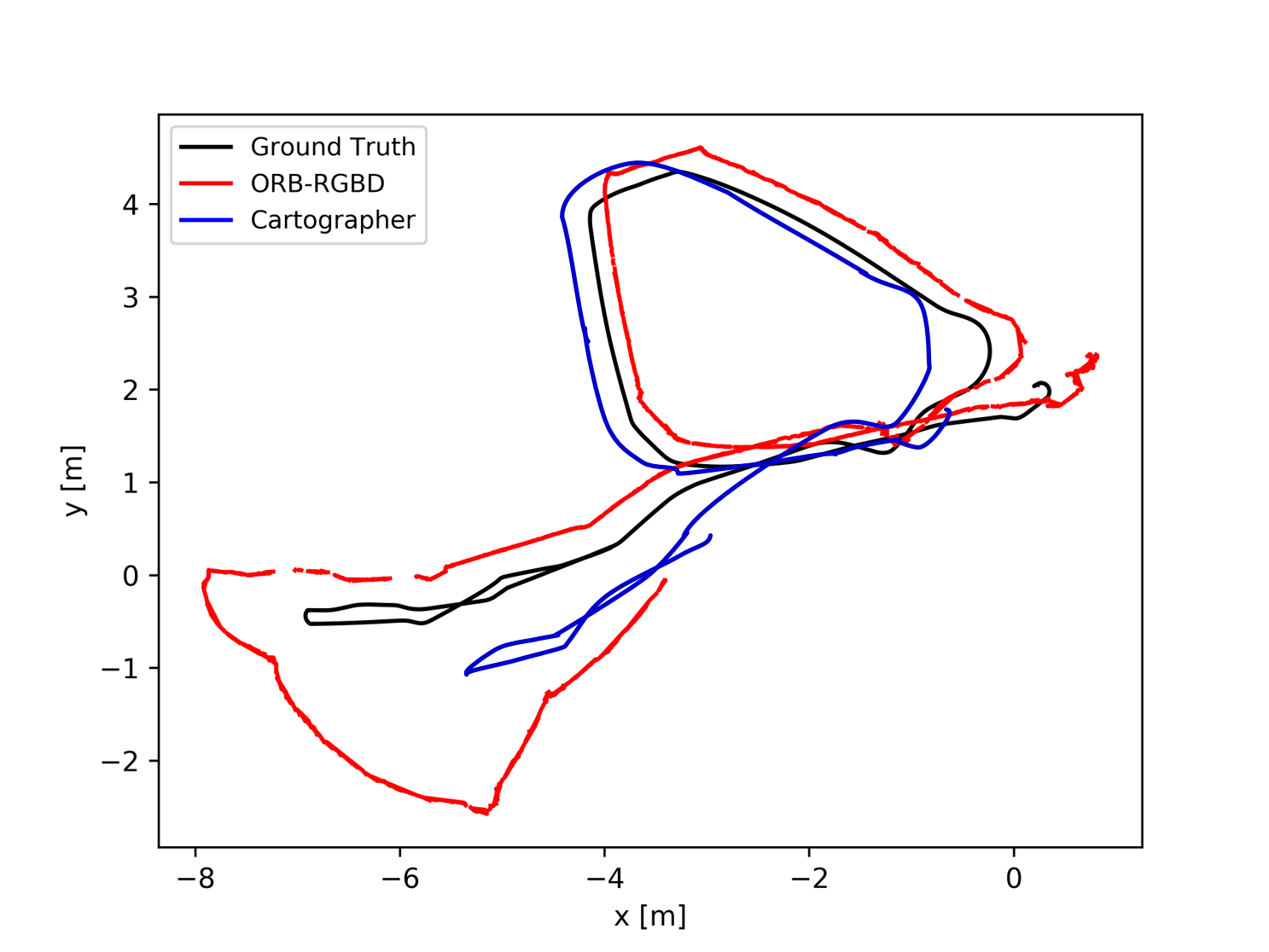}
        \label{fig:traj1}    
    \end{subfigure} 
    \begin{subfigure}[H]{\linewidth}    
        \includegraphics[width=47mm,height=39.5mm]{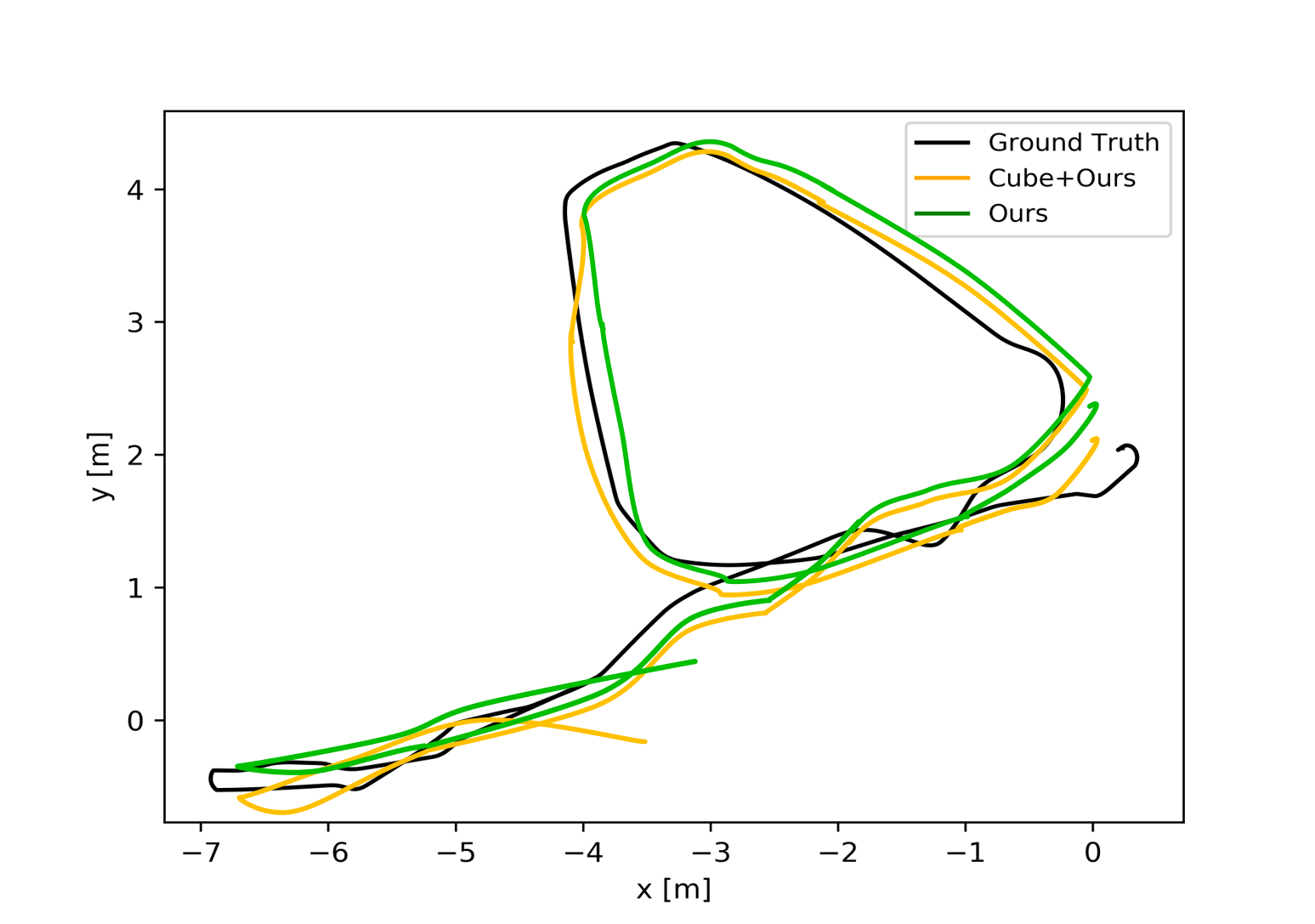}
        \label{fig:traj2}    
    \end{subfigure} 
  \end{multicols}  
    \centering
    \caption{\small Trajectory Plotting}
    \label{fig:traj}
\end{figure}

From above, the result in bottom row in Table \ref{slam-result} is consistent with the plotting, both methods integrating object-landmark can generate more consistent trajectory with ground truth, while "ORB-RGBD" in red \cite{ORB} and "cartographer" in blue \cite{carto} suffer from obvious mismatch during entering into small room with very cluttered environment, even tracking lost may happen sometimes. In addition, more rich front-end 3D model adaptive to the real object shape can achieve a better accuracy over cuboid-only approach, like cuboid orientation estimation is very inaccurate for swivel chair. The 2D occupancy map result from "cartographer" \cite{carto}, with and without landmark constraints are presented below. For post-optimization, the landmark weighting is inversely proportional with the pose covariance.

\begin{figure}[H]   
   \begin{multicols}{2}
    \begin{subfigure}[H]{\linewidth}    
        \includegraphics[width=43mm,height=46.5mm]{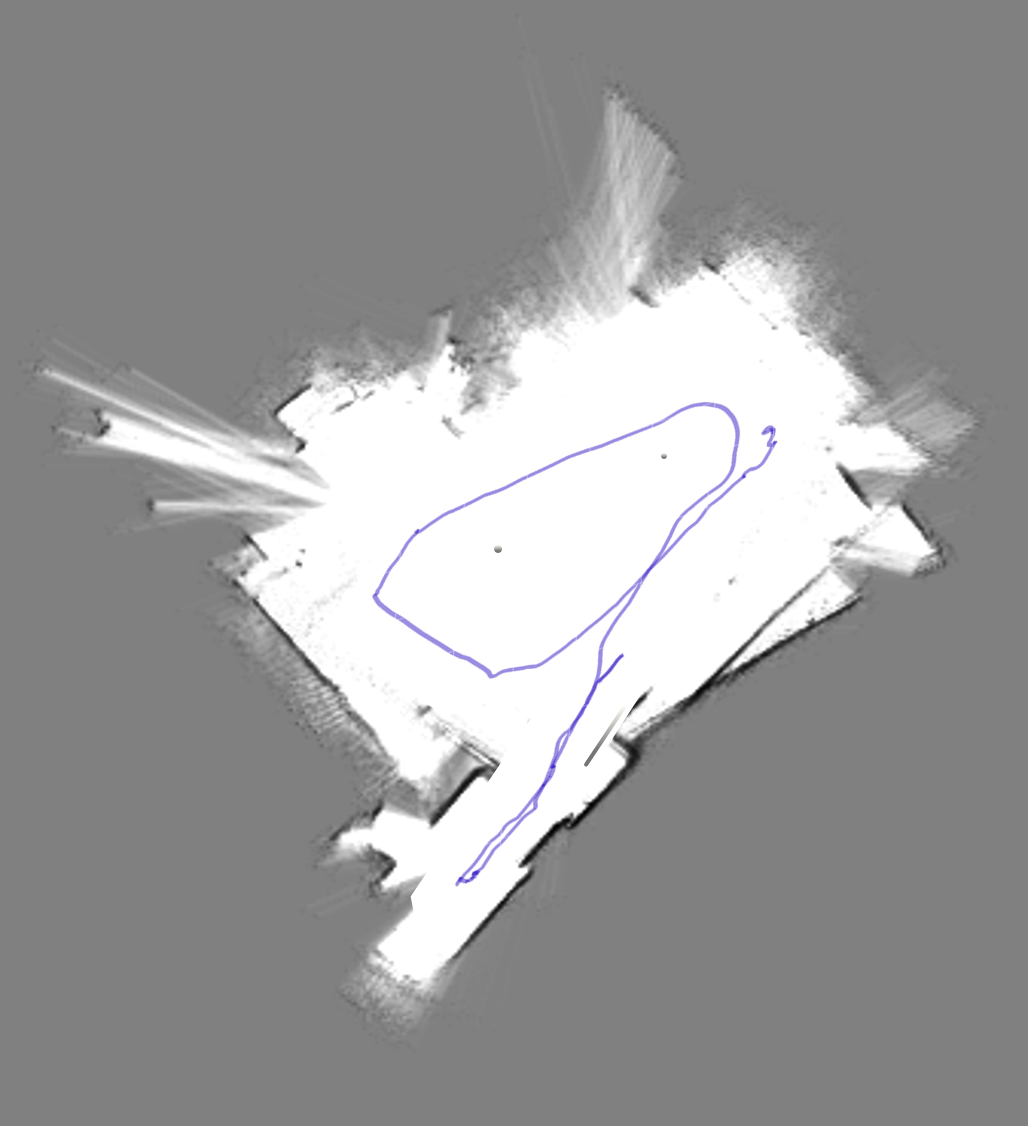}
        \label{fig:badmap}    
    \end{subfigure} 
    \begin{subfigure}[H]{\linewidth}    
        \includegraphics[width=43mm,height=46.5mm]{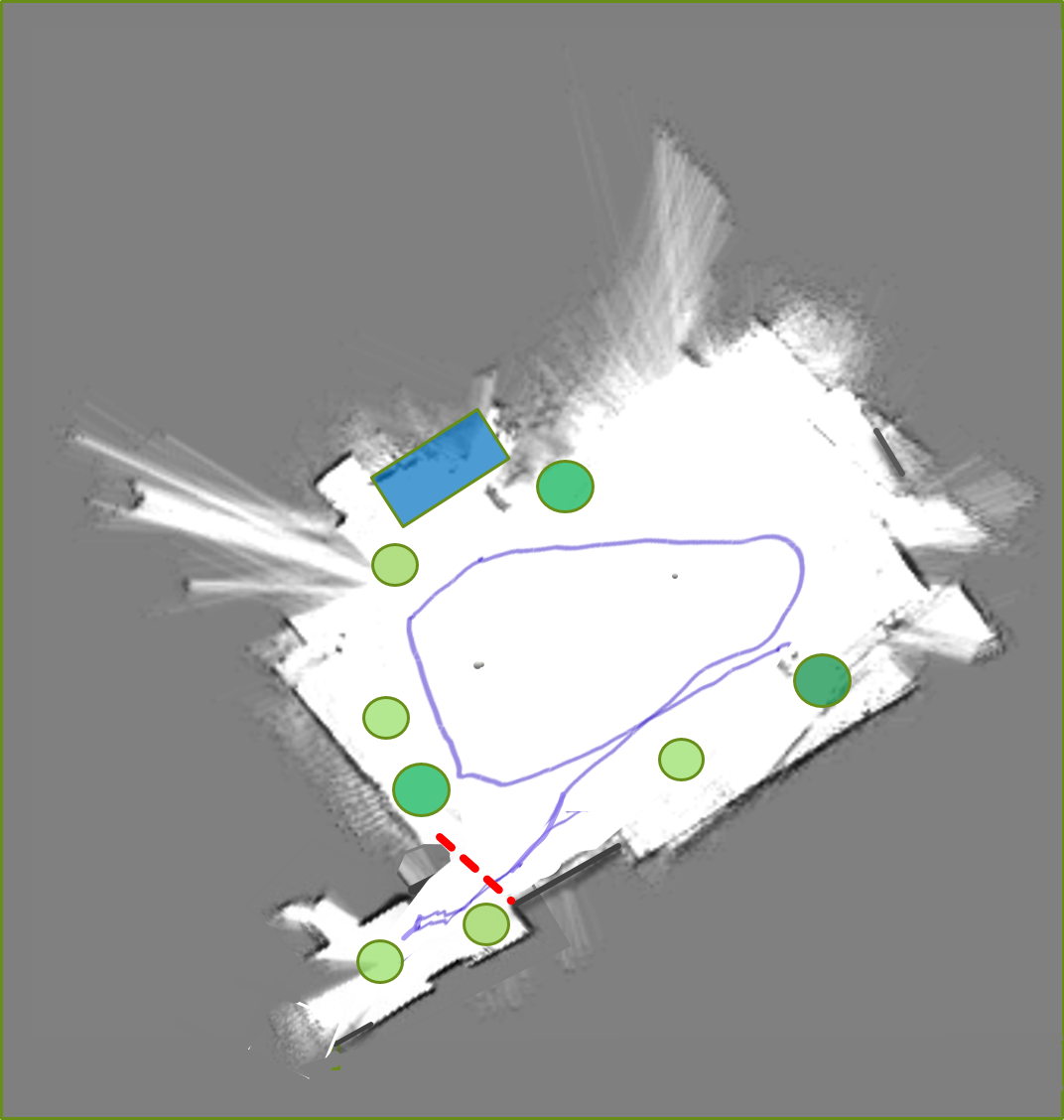}
        \label{fig:landmarkmap}    
    \end{subfigure} 
  \end{multicols}  
    \centering
    \caption{\small Top view of 2D Map with and without object-Landmark }
    \label{fig:map-comp}
\end{figure}

The sub-figure on the right shows layout of detected swivel chair and Sofa on the map. The red dashed line indicates a door detection, which signifies new landmark initialization.

A final 3D demonstration result in "rviz" is presented below. On the right side of big room is non-navigable space separated by a big net over ground as in Figure 11.
   \begin{figure}[H]
     \centering
      \includegraphics[scale = 0.26]{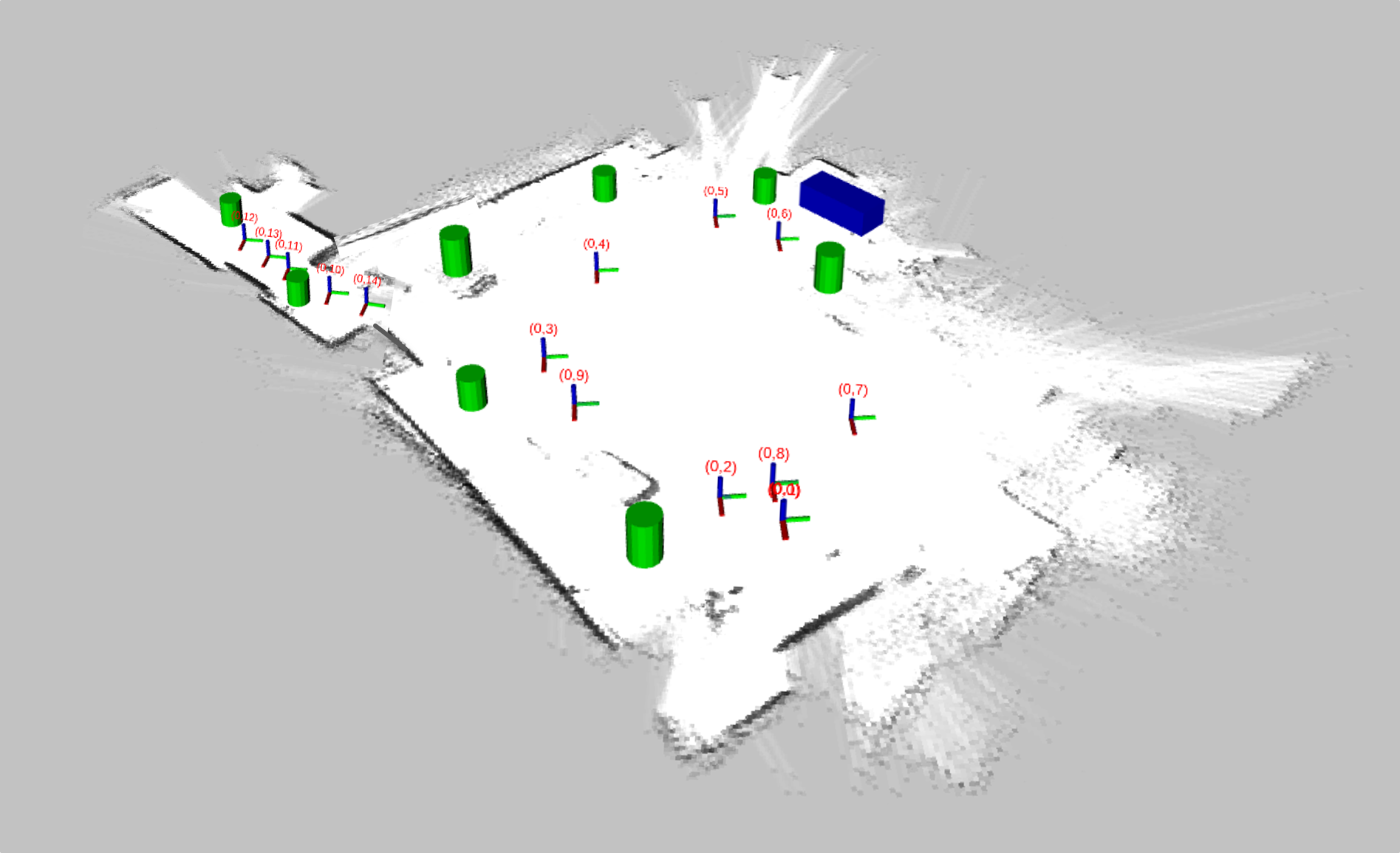} 
      \caption{\small Final Mapping Result Illustration.}
      \label{fig:3dmap}
   \end{figure} 
\section{CONCLUSIONS}

We demonstrate a real-time semantic SLAM framework in this paper, this should be the first time to leverage both camera and 2D Lidar for real SLAM task with unknown data association. The individual sensor merit can complement each other to provide rich perceptual and reliable perception. The system back-end can optimize both the semantic, geometric information jointly in a soft but robust way.    

The system overview diagram is presented at start. Then the detection neural network "Yolov3" along with edge extracting, and a rough object depth mask can generate several initial 2D proposals for 3D model, followed by CRF filtering, to find a best configuration of these proposals. The proposal back-projected to world, along with feature points triangulated to 3D, and odometry pose form a factor graph, will be optimized with unknown data association coordinately, the semantic discrete variables are plugged into the association step, to help search for the best matching heuristically, the semantic information in each frame is encoded into a binary string. 

Finally the whole framework is evaluated on our own made data sequence, collected in a real challenging indoor environment with cluttering and dynamic persons, even further verified by open-source dataset "ICL-NUIM". The performance achieves a good accuracy, and proves the improving of pose and mapping accuracy by using semantic information. The system can provide a compact and reusable map in the end. We further looks forward to extending our work to cope with multi-object tracking, and merged with auto-exploration to realize a more intelligent map building, and efficient maintaining way to bridge the gap in productization.  
\addtolength{\textheight}{-12cm}   

\section*{ACKNOWLEDGMENT}

During work, our colleague, Walker Wang helped us to set up experimental robot platform.

\bibliographystyle{IEEEconf}


\end{document}